%% file: main.tex
\documentclass[10pt,journal,compsoc]{IEEEtran}

\ifCLASSOPTIONcompsoc
\else
  \usepackage{cite}
\fi

\usepackage[utf8]{inputenc}
\input{packages}

\input{macros}

\usepackage[colorlinks=true,linkcolor=blue,urlcolor=blue,citecolor=blue]{hyperref}
\usepackage{graphicx}
\usepackage{color}
\usepackage{subcaption}
\usepackage{amsmath, amsthm, amssymb}
\usepackage{multirow}
\usepackage[square,sort,comma,numbers]{natbib}
\usepackage{pifont}

\begin{document}
%
\title{Design Automation for Efficient \\ Deep Learning Computing}

\author{
Song Han, Han Cai, Ligeng Zhu, Ji Lin, Kuan Wang, Zhijian Liu, Yujun Lin

Massachusetts Institute of Technology

\{songhan, hancai, ligeng, jilin, kuanwang, zhijian, yujunlin\}@mit.edu
}

\markboth{}%
{Shell \MakeLowercase{\textit{et al.}}: Bare Demo of IEEEtran.cls for Computer Society Journals}

\IEEEtitleabstractindextext{%
\begin{abstract}
\input{contents/00_abstract}

\end{abstract}

}

\maketitle

\IEEEdisplaynontitleabstractindextext

%
\IEEEpeerreviewmaketitle

\input{contents/01_intro.tex}

\input{contents/02_neural_architecture_design.tex}
\input{contents/03_pruning.tex}

\input{contents/04_quantization.tex}

\input{contents/05_conclusion.tex}

\newpage

\newpage

\ifCLASSOPTIONcaptionsoff
  \newpage
\fi


\bibliographystyle{IEEEtran}
\bibliography{main}


\end{document}

%% file: packages.tex
\usepackage{color,xcolor}
\usepackage{epsfig}
\usepackage{graphicx}
\usepackage{subfiles}

\usepackage{adjustbox}
\usepackage{array}
\newcolumntype{?}[1]{!{\vrule width #1}}
\usepackage{booktabs}
\usepackage{colortbl}
\usepackage{float,wrapfig}
\usepackage{hhline}
\usepackage{multirow}
\usepackage{subcaption} 
\usepackage[labelfont={bf},labelsep={period},font={small}]{caption}

\let\llncssubparagraph\subparagraph
\let\subparagraph\paragraph
\usepackage[compact]{titlesec}
\let\subparagraph\llncssubparagraph

\usepackage{amsmath,amsfonts,amssymb}
\usepackage{bm}
\usepackage{nicefrac}
\usepackage{microtype}

\usepackage{changepage}
\usepackage{extramarks}
\usepackage{fancyhdr}
\usepackage{lastpage}
\usepackage{setspace}
\usepackage{soul}
\usepackage{xspace}

\usepackage{url}

\usepackage{algorithm}
\usepackage{algpseudocode}
\usepackage{enumerate}
\usepackage{footnote}

%% file: macros.tex
\newcommand{\action}{\textit{a}}

\newcommand{\ours}{AMC (ours)}
\newcommand{\AMC}{AMC\xspace}

\newcolumntype{L}[1]{>{\raggedright\let\newline\\\arraybackslash\hspace{0pt}}m{#1}}
\newcolumntype{C}[1]{>{\centering\let\newline\\\arraybackslash\hspace{0pt}}m{#1}}
\newcolumntype{R}[1]{>{\raggedleft\let\newline\\\arraybackslash\hspace{0pt}}m{#1}}

\newcommand{\fig}[1]{Figure~\ref{#1}}
\newcommand{\tbl}[1]{Table~\ref{#1}}

\newcommand{\ignore}[1]{}

\makeatletter
\DeclareRobustCommand\onedot{\futurelet\@let@token\@onedot}
\def\@onedot{\ifx\@let@token.\else.\null\fi\xspace}

\def\eg{\emph{e.g}\onedot} 
\def\ie{\emph{i.e}\onedot}

\makeatother

\definecolor{MyDarkBlue}{rgb}{0,0.08,1}
\definecolor{MyDarkGreen}{rgb}{0.02,0.6,0.02}
\definecolor{MyDarkRed}{rgb}{0.8,0.02,0.02}
\definecolor{MyDarkOrange}{rgb}{0.40,0.2,0.02}
\definecolor{MyPurple}{RGB}{111,0,255}
\definecolor{MyRed}{rgb}{1.0,0.0,0.0}
\definecolor{MyGold}{rgb}{0.75,0.6,0.12}
\definecolor{MyDarkgray}{rgb}{0.66, 0.66, 0.66}

\algnewcommand{\LeftComment}[1]{\Statex \(\triangleright\) #1}

%% file: contents/00_abstract.tex
Efficient deep learning computing requires algorithm and hardware co-design to enable specialization: we usually need to change the algorithm to reduce memory footprint and improve energy efficiency. However, the extra degree of freedom from the algorithm makes the design space much larger: it's not only about designing the hardware but also about how to tweak the algorithm to best fit the hardware. 
Human engineers can hardly exhaust the design space by heuristics. It’s labor consuming and sub-optimal. We propose design automation techniques for efficient neural networks. We investigate automatically designing specialized fast models, auto channel pruning, and auto mixed-precision quantization. We demonstrate such learning-based, automated design achieves superior performance and efficiency than rule-based human design. 
Moreover, we shorten the design cycle by 200$\times$ than previous work, so that we can afford to design \emph{specialized} neural network models for different hardware platforms.

%% file: contents/01_intro.tex
\section{Introduction}
Algorithm and hardware co-design plays an important role in efficient deep learning computing. Unlike optimizing on the SPEC2006 benchmark when we can treat the algorithm as a black box, there's plenty of room at the algorithm level that can improve the hardware efficiency of deep learning. We should open the box and explore model optimization techniques. The benefit usually comes from memory saving and locality improvement. For example, model compression techniques \cite{han2015deep} including pruning and quantization can drastically reduce the memory footprint and save energy consumption. Another example is small model design. SqueezeNet \cite{iandola2016squeezenet} and MobileNet \cite{howard2017mobilenets} have only 4.8MB/4.2MB of model size, which can fit on-chip SRAM and improve the locality.

However, efficient model design and compression have a large design space. Many different neural network architectures can lead to similar accuracy, but drastically different hardware efficiency. This is difficult to exhaust by rule-based heuristics, since there is a shortage of deep learning and hardware experts to hand-tune each model to make it run fast. It's demanding to systematically study how to design efficient neural network with hardware constraints. We propose hardware-centric AutoML techniques that can automatically design neural networks that are hardware efficient \cite{cai2018proxylessnas, he2018amc,wang2018haq}. Such joint optimization is systematic and can transfer well between tasks. It requires fewer engineer efforts while designing better neural networks at low cost.

\begin{figure*}[t]
	\centering
	\includegraphics[width=1\linewidth]{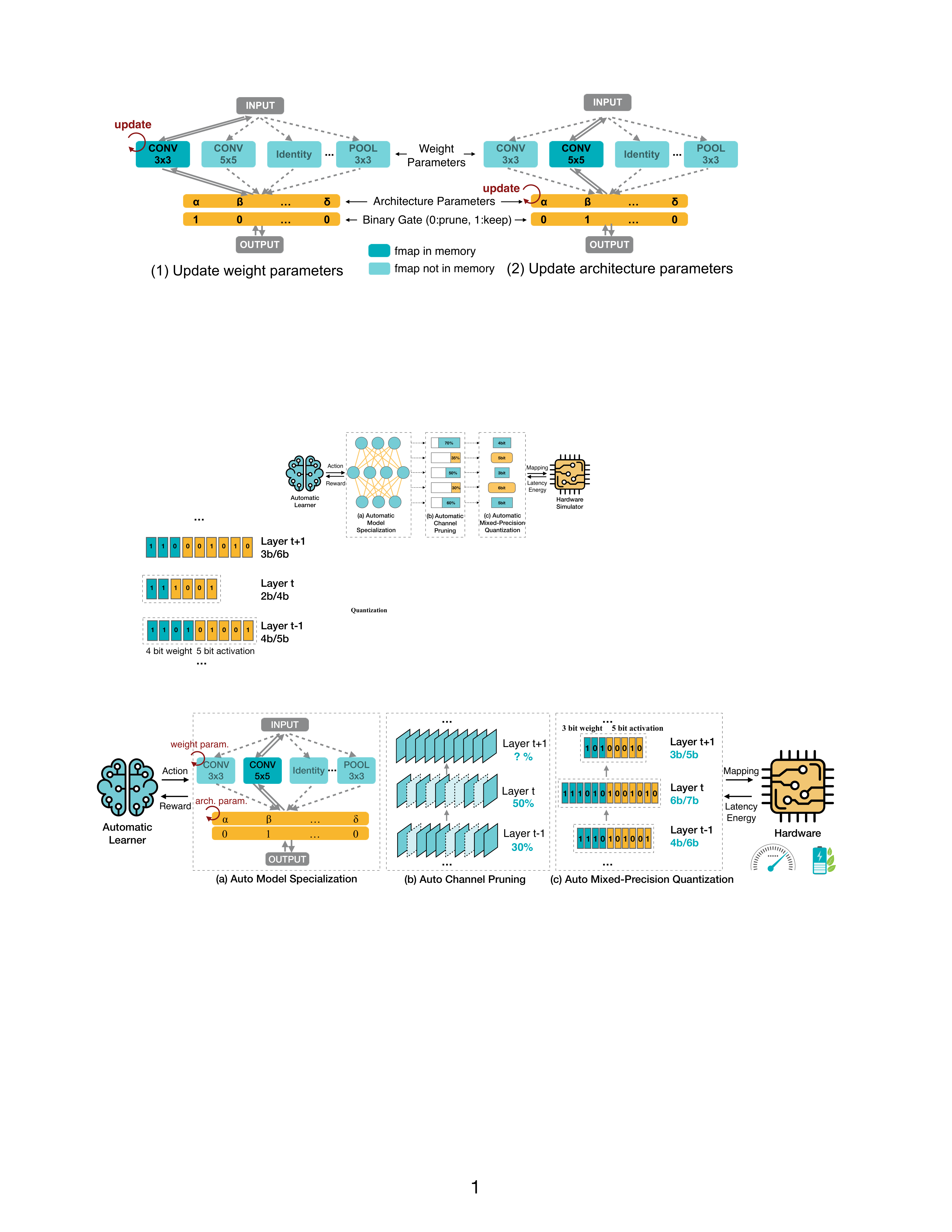}
	\caption{Design automation for model specialization, channel pruning and mixed-precision quantization.}
	\label{fig:method}
\end{figure*}

We explore three aspects of neural network design automation (Figure~\ref{fig:method}): auto design specialized model, auto channel pruning, and auto mixed-precision quantization. Each aspect is summarized as follows. 

There is plenty of specialized hardware for neural networks, but little research has been done for specialized neural network architecture for a given hardware architecture (the \emph{reverse} specialization). There are several advantages for a specialized model: it can fully utilize the parallelism of the target hardware (e.g. fitting the channel size with the PE size). Besides, a specialized model can fully utilize the on-chip buffer and improve locality and reuse. Specialization can also match the model's computation intensity with the hardware's roofline model. However, designing a specialized neural network architecture used to be difficult. First, there are limited heuristics. Second, the  computation cost used to be prohibitive: even searching a model on CIFAR-10 dataset takes $10^4$ GPU hours \cite{zoph2016neural, zoph2017learning}. We cut the search cost by two orders of magnitude (actually more than that, since we directly search on ImageNet). The search cost is reduced by two techniques: path-level pruning and path-level binarization, which saves GPU hours and GPU memory. Cutting the search cost enables us to design specialized the model on the target task and target hardware. On the mobile phone, our searched model \cite{cai2018proxylessnas} runs 1.8$\times$ faster than the best human designed model \cite{sandler2018mobilenetv2}. 

After designing a specialized model, compression and pruning is an effective technique to further reduce the memory footprint~\cite{han2015deep}. Conventional model compression techniques rely on hand-crafted heuristics and \emph{rule-based} policies that require domain experts to explore the large design space. We propose an automated design flow that leverages reinforcement learning to give the best model compression policy. This \emph{learning-based} compression policy outperforms conventional \emph{rule-based} compression policy by having a higher compression ratio, better preserving the accuracy and freeing human labor. We applied this automated, push-the-button compression pipeline to MobileNet and achieved \textbf{1.81$\times$} speedup of measured inference latency on an Android phone and \textbf{1.43$\times$} speedup on the Titan XP GPU, with only 0.1\% loss of ImageNet Top-1 accuracy.

The last step is automatic mixed-precision quantization. Emergent DNN hardware accelerators begin to support \emph{flexible bitwidth} (1-8 bits), which raises a great challenge to find the optimal bitwidth for each layer: it requires domain experts to explore the vast design space trading off among accuracy, latency, energy, and model size. Conventional quantization algorithm ignores the different hardware architectures and quantizes all the layers in a uniform way. We introduce the automated design flow of model quantization, and we take the hardware accelerator's feedback in the design loop. Our framework can specialize the quantization policy for different hardware architectures. It can effectively reduce the latency by \textbf{1.4-1.95$\times$} and the energy consumption by \textbf{1.9$\times$} with negligible loss of accuracy compared with the fixed bitwidth (8 bits) quantization. 


%% file: contents/02_neural_architecture_design.tex
\section{Automated Model Specialization}

In order to fully utilize the hardware resource, we propose to search a specialized CNN architecture for the given hardware. The model is compact and runs fast. 
We start with a large design space (Figure~\ref{fig:method}(a)) that includes many candidate paths to \emph{learn} which is the best one by gradient descent, rather than hand-picking with rule-based heuristics. Instead of just learning the weight parameter, we jointly learn the architecture parameter (shown in red in Figure~\ref{fig:method}(a)). The architecture parameter is the probability of choosing each path. 
The search space for each block $i$ consists of many choices:
\begin{itemize}
    \item \texttt{ConvOp}: mobile inverted bottleneck conv~\cite{sandler2018mobilenetv2} with various kernel sizes and expansion ratios
    \begin{itemize}
        \item Kernel size: \{3$\times$3, 5$\times$5, 7$\times$7\}
        \item Expansion ratio: \{3, 6\} 
    \end{itemize}
    \item \texttt{ZeroOp}: if \texttt{ZeroOp} is chosen at $i^\text{th}$ block, it means the block is skipped.
\end{itemize}

Therefore, the number of possible architectures in the design space is $[\underbrace{(3\times2)}_\texttt{ConvOp} + \underbrace{1}_\texttt{ZeroOp}]^{N} = 7^{N}$ where $N$ is the number of blocks (21 in our experiments). 

Given the vast design space, it is infeasible to rely on domain experts to manually design the CNN model for each hardware platform. So we need to employ automatic architecture design techniques. 

However, early reinforcement learning-based \cite{zoph2016neural,zoph2017learning} NAS methods are very expensive to run (\eg, $10^4$ GPU hours) since they need to iteratively sample an architecture, train it from scratch and update the meta-controller. It typically requires tens of thousands of networks to be trained to find a good neural network architecture.

We adopt a different approach to improve the efficiency of model specialization \cite{cai2018proxylessnas}. We first build a super network that comprises all candidate architectures. Concretely, it has a similar structure to a CNN model in the design space except that each specific operation is replaced with a mixed operation that has $n$ parallel paths. Each path in a mixed operation corresponds to a candidate operation $o_i(\cdot)$, and we introduce an architecture parameter $\alpha_i$ to each path to learn which paths are redundant and thereby can be pruned (i.e. path-level prunning). 

In the forward step, to save GPU memory, we allow only one candidate path to actively reside in the GPU memory. This is achieved by hard-thresholding the probability of each candidate path to either 0 or 1 (\ie, path-level binarization). As such the output of a mixed operation is given as 
\begin{align}\label{eq:mix_op}
    x_{l} &= \sum_i g_i o_i (x_{l-1}) 
\end{align}
where $g_i$ is sampled according to the multinomial distribution derived from the architecture parameters, \ie,  \{$p_i = \text{softmax}(\alpha_i; \alpha) = \exp(\alpha_i) / \sum_i \exp(\alpha_i)$\}.

In the backward step, we update the weight parameters of active paths using standard gradient descent. Since the architecture parameters are not directly involved in the computational graph (Eq.~\ref{eq:mix_op}), we use the gradient w.r.t. binary gates to update the corresponding architecture parameters:
\begin{align} 
\begin{split}
    \frac{\partial L}{\partial \alpha_i} 
    & = \sum_{j = 1} \frac{\partial L}{\partial p_j} \frac{\partial p_j}{\partial \alpha_i} 
    \approx \sum_{j = 1} \frac{\partial L}{\partial g_j} \frac{\partial p_j}{\partial \alpha_i} .
\end{split} \nonumber
\end{align}

In order to specialize the model for hardware, we need to take the latency running on the hardware as a design reward. However, directly measuring the inference latency suffer from (i) slow (ii) high variance due to different battery condition and thermal throttling (iii) latency is non-differentiable and can't be directly optimized. To address these, we present our latency prediction model and hardware-aware loss. 

To build the latency model we pre-compute the latency of each operator with all possible inputs. During search we query the lookup table during the searching process 
. The overall latency of $i^{th}$ block is the weighted sum of the latency of each operator.


\begin{table}[t]
    \small\centering
    \begin{tabular}{lccc}
    \toprule
	Model & Top-1 & Top-5 & GPU Latency \\
	\midrule
	MobileNet-V2 \citep{sandler2018mobilenetv2} & 72.0 & 91.0 & 6.1 ms \\
	ResNet-34 \citep{he2016deep} & 73.3 & 91.4 & 8.0 ms \\
	\midrule
	NASNet-A \citep{zoph2017learning} & 74.0 & 91.3 & 38.3 ms \\
	MnasNet \citep{tan2018mnasnet} & 74.0 & 91.8 & 6.1 ms \\
	\midrule
	Specialized model for GPU & \textbf{75.1} & \textbf{92.5} & \textbf{5.1 ms}  \\
    \bottomrule
    \end{tabular}
	\caption{ImageNet Accuracy (\%) and GPU latency (Tesla V100).}
	\label{tab:proxyless_imagenet_gpu}
\end{table}

\begin{align}\label{eq:latency_sum}
\begin{split}
    \mathbb{E}[\mathrm{LAT_i}] = \; & \alpha \times F(\mathrm{mb3\_3x3}) + \\
    & \beta  \times F(\mathrm{mb3\_5x5}) + \\
    & \sigma \times F(\mathrm{identity}) + \\
    & ...... \\
    &\zeta  \times F(\mathrm{mb6\_7x7})\\
    \mathbb{E}[\mathrm{LAT}] &= \sum_{i}^{N} \mathbb{E}[\mathrm{LAT_i}]
\end{split}
\end{align}

Then we combine the latency and training loss (e.g. cross-entropy loss) using the following formula
\begin{equation}\label{eq:latency_loss}
    \mathcal{L} = \mathcal{L}_\mathrm{CE} \times \alpha 
    \log\left(
        \frac{
            \mathbb{E}[\mathrm{LAT}]
        }{
            \mathrm{LAT}_{\mathrm{ref}}
        }
    \right)^{\beta},
\end{equation}
where $\alpha$ and $\beta$ are hyper-parameters controlling the trade-off between accuracy and latency and $\mathrm{LAT}_{\mathrm{ref}}$ is the target latency. Note our formulation not only provides a fast estimation of the searched model but also makes the search process fully differentiable. 

\begin{figure}[t]
    \centering
    \includegraphics[width=0.7\linewidth]{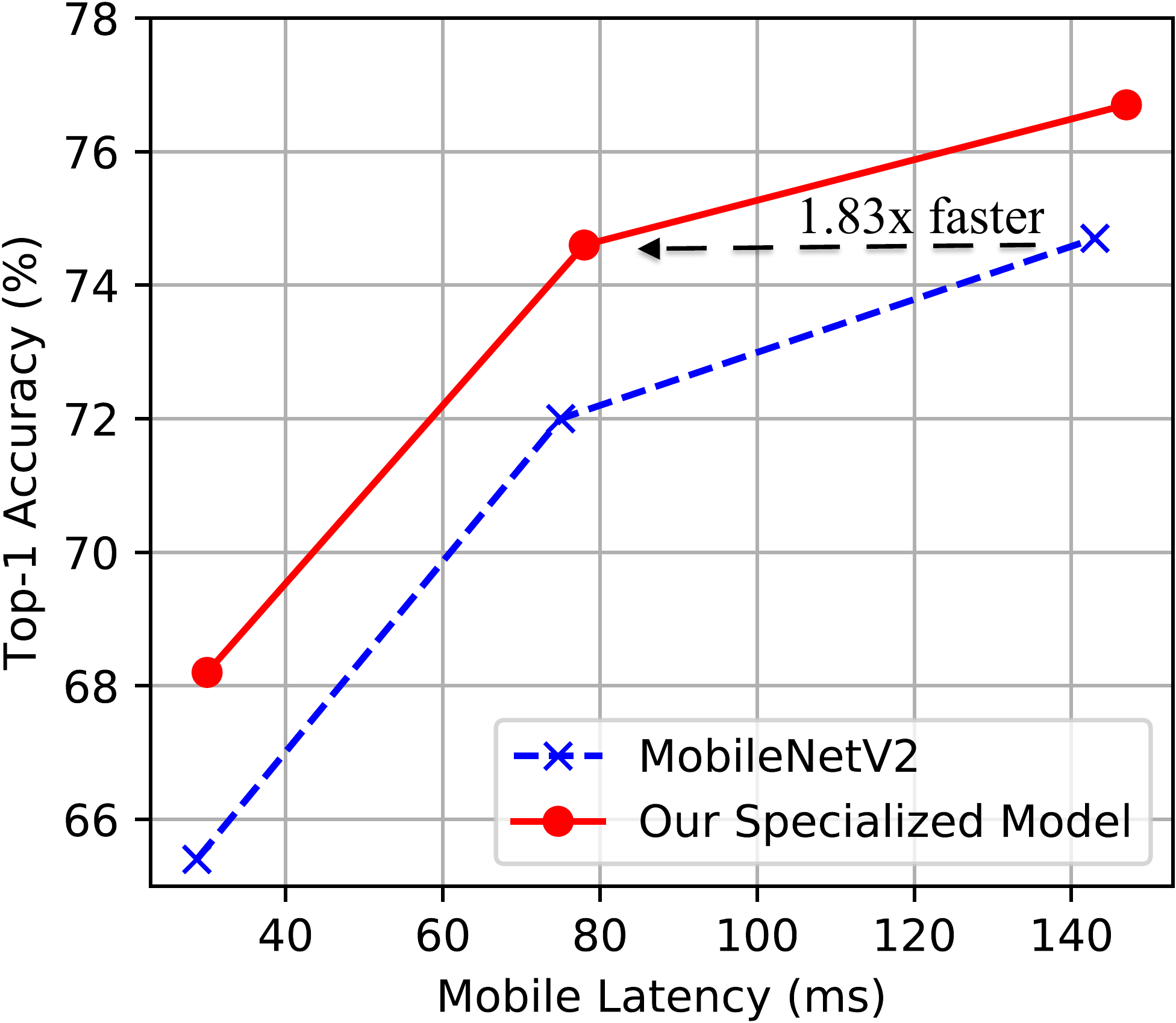}
    \caption{AI automatically designed specialized model consistently outperforms human designed MobileNetV2 under various latency settings.}
    \label{fig:proxyless_mobile_rescale}
\end{figure}

We demonstrate the effectiveness of our model specialization on ImageNet dataset with CPU (Xeon E5-2640 v4), GPU (Tesla V100) and mobile phone (Google Pixel-1).  
We first search for a specialized CNN model for the mobile phone (Figure~\ref{fig:proxyless_mobile_rescale}). Compared to MobileNet-V2 (the state-of-the-art human engineered architecture), our model improves the top-1 accuracy by 2.6\% while maintaining a similar latency. Under the same level of top-1 accuracy (around 74.6\%), \textbf{MobileNet-V2 has 143ms latency while ours has only 78ms (1.83$\times$ faster)}. Compared with the state-of-the-art auto designed model, MnasNet~\citep{tan2018mnasnet}, our model can achieve 0.6\% higher top-1 accuracy with slightly lower mobile latency. More remarkably, \textbf{we reduced the search cost by 200$\times$, from 40,000 GPU hours to only 200 GPU hours}.

Table~\ref{tab:proxyless_imagenet_gpu} reports the speedup on GPU. our method achieved superior performances compared to both human-designed and automatically searched architectures. Compared general purpose models, our specialized model improves the top-1 accuracy by 1.1\% - 3.1\% while being 1.2$\times$-7.5$\times$ faster. Table~\ref{tab:proxyless_imagenet_discussion} compares the specialized models on CPU/GPU/Mobile. As expected, models specialized for GPU do not run fast on CPU and mobile phone, vice versa. 
It is essential to learn \emph{specialized} neural networks to cater for different hardware.

Our automated design flow designed CNN architectures that were long dismissed as being too inefficient — but in fact, they are very efficient. For instance, engineers have essentially stopped using 7$\times$7 filters, because they’re computationally more expensive than multiple, smaller filters (one 7$\times$7 layer has the same receptive field than three 3$\times$3 layers, but bears 49 weights rather than 27.). However, our AI designed model found that using 7$\times$7 filter is very efficient on GPUs. That’s because GPUs have high parallelization, and invoking a large kernel call is more efficient than invoking multiple small kernel calls. This design choice goes against previous human thinking. The larger the search space, the more unknown things you can find. You don’t know if something will be better than the past human experience. Let the automated design tool figure it out \cite{mitnews}.

\begin{table}[t]
    \renewcommand*{\arraystretch}{1.4}
    \setlength{\tabcolsep}{6pt}
	\small\centering
	\begin{tabular}{lcccc}
		\toprule
		Model & Top-1 & GPU & CPU & Mobile \\
		\midrule
		Specialized for GPU  & 75.1 & \cellcolor{red!15}{\textbf{5.1ms}} & 204.9ms & 124ms \\
		Specialized for CPU & 75.3 & 7.4ms & \cellcolor{red!15}{\textbf{138.7ms}} & 116ms \\
		Specialized for Mobile & 74.6 & 7.2ms & 164.1ms & \cellcolor{red!15}{\textbf{78ms}} \\
		\bottomrule
	\end{tabular}
	\caption{Hardware prefers specialized models. Models optimized for GPU does not run fast on CPU and mobile phone, vice versa. Our method provides an efficient solution to search a specialized neural network architecture for a target hardware architecture, while cutting down the search cost by 200$\times$ compared with state-of-the-arts \citep{zoph2016neural, tan2018mnasnet}. }\label{tab:proxyless_imagenet_discussion}
\end{table}

%% file: contents/03_pruning.tex
\section{Automated Channel Pruning}


Pruning \cite{han2015learning} is widely used in model compression and acceleration. It is very important to find the optimal sparsity for each layer during pruning. Pruning too much will hurt accuracy; too less will not achieve high compression ratio. This used to be manually determined in previous studies \cite{han2015deep}. Our goal is to automatically find out the effective \emph{sparsity} for each layer. We train an reinforcement learning agent to predict the best sparsity for a give hardware \cite{he2018amc}. We evaluate the accuracy and FLOPs after pruning. Then we update the agent by encouraging smaller, faster and more accurate models.

\begin{table*}[t]
\small\centering 
\begin{tabular}{c|c|cc|cc|ccc}
\toprule
\multirow{2}{*}{} & \multirow{2}{*}{\begin{tabular}[c]{@{}c@{}}Million\\ MAC\end{tabular}} & \multirow{2}{*}{\begin{tabular}[c]{@{}c@{}}Top-1\\ Acc.\end{tabular}} & \multirow{2}{*}{\begin{tabular}[c]{@{}c@{}}Top-5\\ Acc.\end{tabular}} & \multicolumn{2}{c|}{GPU} & \multicolumn{3}{c}{Android} \\
 &  &  &  & \begin{tabular}[c]{@{}c@{}}Latency \\\end{tabular} & \begin{tabular}[c]{@{}c@{}}Speed \\\end{tabular} & 
\begin{tabular}[c]{@{}c@{}}Latency \\\end{tabular} & \begin{tabular}[c]{@{}c@{}}Speed \\\end{tabular} & \begin{tabular}[c]{@{}c@{}}Memory \\\end{tabular} \\

\midrule

\begin{tabular}[c]{@{}c@{}}100\%\\ MobileNet\end{tabular} & 569 & 70.6\% & 89.5\% & 0.46ms & 2191 fps &  123.3ms & 8.1 fps & 20.1MB \\ 
\begin{tabular}[c]{@{}c@{}}75\%\\ MobileNet\end{tabular} & 325 & 68.4\% & 88.2\% & 0.34ms & 2944 fps &  72.3ms &13.8 fps & 14.8MB \\

\midrule

\begin{tabular}[c]{@{}c@{}}\textbf{AMC} \\ (50\% FLOPs)\\ \end{tabular} & 285 & 70.5\% & 89.3\% & 0.32ms & \begin{tabular}[c]{@{}c@{}}\textbf{3127 fps}\\ (\textbf{1.43}$\times$)\end{tabular} &  68.3ms & \begin{tabular}[c]{@{}c@{}}\textbf{14.6 fps}\\ (\textbf{1.81}$\times$)\end{tabular} & 14.3MB \\ 
\begin{tabular}[c]{@{}c@{}}\textbf{AMC} \\ (50\% Latency) \end{tabular} & 272 & 70.2\% & 89.2\% & 0.30ms & \begin{tabular}[c]{@{}c@{}}\textbf{3350 fps}\\ (\textbf{1.53}$\times$)\end{tabular} & 63.3ms & \begin{tabular}[c]{@{}c@{}}\textbf{16.0 fps}\\ (\textbf{1.95}$\times$)\end{tabular}  & 13.2MB \\
\bottomrule
\end{tabular}  
\caption{AMC speeds up MobileNet. On Google Pixel-1 CPU, \AMC achieves  1.95$\times$ measured speedup with batch size one, while saving the memory by 34\%. On NVIDIA Titan XP GPU, \AMC achieves 1.53$\times$ speedup with batch size of 50.
}
\label{tab:amc_android}
\end{table*}

Our automatic model compression (AMC) leverages reinforcement learning to efficiently search the pruning ratio (Figure~\ref{fig:method}(b)). The reinforcement learning agent receives an embedding state $s_t$ of layer $L_t$ from the environment and then outputs a sparsity ratio as action $\action_t$. The layer is compressed with $\action_t$ (rounded to the nearest feasible fraction). Then the agent moves to the next layer $L_{t+1}$, and receives state $s_{t+1}$.
After finishing the final layer $L_T$, the reward accuracy is evaluated on the validation set and returned to the agent. 
With our framework, we are able to push the expert-tuned limit of fine-grained model pruning. For ResNet-50 on ImageNet, we can push the compression ratio from 3.4$\times$ to 5$\mathbf{\times}$ without loss of accuracy. With further investigation, we find that AMC automatically learns to prune 3$\times$3 convolution kernels harder than 1$\times$1 kernels, which is similar to human heuristics since the latter is less redundant.


\begin{table}[t]
\centering
\resizebox{1.0\linewidth}{!}{
\begin{tabular}{c|c|c|c}
\hline
 & Policy & FLOPs & $\Delta$Acc (\%) \\ \hline
\multirow{4}{*}{MobileNet-V1} 
& uniform (0.75-224)~\cite{howard2017mobilenets}  &  56\%  & -2.5     \\ 
\cline{2-4} 
& \textbf{\ours}& 50\% & \textbf{-0.4}  \\
\cline{2-4} 
& uniform (0.75-192)~\cite{howard2017mobilenets} & 41\% & -3.7 \\
\cline{2-4} 
& \textbf{\ours}& 40\% & \textbf{-1.7}     \\

\hline \hline
\multirow{2}{*}{\ MobileNet-V2\ } 
& uniform (0.75-224)~\cite{sandler2018mobilenetv2}  &  
\multirow{2}{*}{\begin{tabular}[c]{@{}c@{}}70\% \end{tabular}}  
& -2.0     \\ 
\cline{2-2}  \cline{4-4} 
& \textbf{\ours}& & \textbf{-1.0}  \\
\hline
\end{tabular}
}
\caption{Learning-based automated model compression (AMC) outperforms rule-based model compression. Rule-based heuristics are suboptimal.}
\label{tab:amc_vggde}
\end{table}

We also compare AMC with heuristic-based channel reduction method on modern efficient neural networks MobileNet~\cite{howard2017mobilenets} and MobileNet-V2~\cite{sandler2018mobilenetv2} (Table \ref{tab:amc_vggde}). Since the networks are already very compact, it is convincing to compress these nets. The easiest way to reduce the channels of a model is to use uniform channel shrinkage, \ie use a width multiplier to uniformly reduce the channels of each layer with a fixed ratio. Both MobileNet and MobileNet-V2 present the performance of different multiplier and input sizes, and we compare our pruned result with models of same computations. The format are denoted as \emph{uniform (depth multiplier - input size)}. We can find that our method consistently outperforms the uniform baselines. Even for the current state-of-the-art efficient model design MobileNet-V2, AMC can still improve its accuracy by $1.0\%$ at the same computation.

Mobile inference acceleration has drawn people's attention in recent years. 
Not only can AMC optimize FLOPs and model size, it can also optimize the inference latency. For all mobile inference experiments, we use TensorFlow Lite framework for timing evaluation.
Our experiment platform is Google Pixel 1. Models pruned to 0.5$\times$ FLOPs and 0.5$\times$ inference time are shown in Table~\ref{tab:amc_android}.
For 0.5$\times$ FLOPs setting, we achieve \textbf{1.81}$\times$ speed up on a Google Pixel 1 phone. For 0.5$\times$ FLOPs setting, we accurately achieve \textbf{1.95}$\times$ speed up, which is very close to actual 2$\times$ target, showing that AMC can directly optimize inference time and achieve accurate speed up ratio.
On GPUs, we also achieve up to 1.5$\times$ speedup, which is less than mobile phone but still significant on an already very compact model. The less speedup is because a GPU has higher degree of parallelism than a mobile phone.

%% file: contents/04_quantization.tex
\section{Automated Mixed-Precision Quantization}

Conventional quantization methods quantize each layer of the model to the same precision. Mixed-precision quantization is more flexible but suffer from a huge design space that's difficult to explore. Meanwhile, as demonstrated in \tbl{tbl:teaser}, the quantization solution optimized on one hardware might not be optimal on the other, which raises the demand for \emph{specialized} policies for different hardware architectures and further increase the design space. Assuming the bitwidth is between 1 to 8 for both weights and activations, then each layer has $8^2$ choices. If we have $M$ different neural network models, each with $N$ layers, on $H$ different hardware platforms, there are in total $O(H \times M \times 8^{2N})$ possible solutions. Rather than using rule-based heuristics, we propose an automated design flow to quantize different layer with mixed precision.  
Our hardware-aware automatic quantization (HAQ)~\cite{wang2018haq} models the quantization task as a reinforcement learning problem. We use the actor-critic model to give the quantization policy (\#bits per layer) (Figure~\ref{fig:method}(c)). The goal is not only high accuracy but also low energy and low latency. 
 
An intuitive reward can be FLOPs or the model size. However, these proxy signals are indirect. They do not translate to latency or energy improvement.  Cache locality, number of kernel calls, memory bandwidth all matters. Instead, we use direct latency and energy feedback from the hardware simulator. Such feedback enables our RL agent to learn the hardware characteristics for different layers: \eg, vanilla convolution has more data reuse and locality, while depthwise convolution has less reuse and worse locality, which makes it memory bounded.


\input{figures/haq/teaser_tables}

\input{figures/haq/bismo_latency_v1_figures.tex}

\input{figures/haq/bismo_latency_tables.tex}

In real-world applications, we have limited resource budgets (\ie, latency, energy, and model size). We would like to find the quantization policy with the best performance given the resource constraint. We encourage our agent to meet the computation budget by limiting the action space. After our RL agent gives actions $\{a_k\}$ to all layers, we measure the amount of resources that will be used by the quantized model. The feedback is directly obtained from the hardware simulator. If the current policy exceeds our resource budget (on latency, energy or model size), we will sequentially decrease the bitwidth of each layer until the constraint is finally satisfied.

We applied HAQ to three different hardware architectures to show the importance of specialized quantization policy. 
Inferencing on edge devices and cloud severs can be quite different, since (1) the batch size on the cloud servers are larger (2) the edge devices are usually limited to low computation resources and memory bandwidth. We use embedded FPGA Xilinx Zynq-7020 as our edge device, and server FPGA Xilinx VU9P as our cloud device to compare the specialized quantization policies.    

Compared to fixed 8-bit quantization (PACT~\cite{Choi:2018uw}), our automated quantization consistently achieved better accuracy under the same latency (see \tbl{tbl:bismo_latency}). With similar accuracy, HAQ can reduce the latency by 1.4-1.95$\times$ compared with the baseline.

Our agent gave specialized quantization policy for edge and cloud accelerators (\fig{fig:bismo_v1}). The policy is quite different on different hardware.
For the activations, the depthwise convolution layers are assigned much less bitwidth than the pointwise layers on the edge device; while on the cloud device, the bitwidth of these two types of layers are similar to each other. For weights, the bitwidth of these types of layers are nearly the same on the edge; while on the cloud, the depthwise convolution layers are assigned much more bitwidth than the pointwise convolution layers.

We interpret the quantization policy's difference between edge and cloud by the roofline model. 
Operation intensity is defined as operations (MACs in neural networks) per DRAM byte accessed. A lower operation intensity indicates that the model suffers more from the memory access. The bottom of \fig{fig:bismo_v1} shows the operation intensity (OPs per byte) of convolution layers in the MobileNet-V1. On edge accelerator, which has much less memory bandwidth, our RL agent allocates \emph{fewer} activation bits to the depthwise convolutions since the activations dominates the memory access. On cloud accelerator, which has more memory bandwidth, our agent allocates \emph{more} bits to the depthwise convolutions and allocates \emph{fewer} bits to the pointwise convolutions to prevent it from being computation bounded. \fig{fig:roofline} shows the roofline model before and after HAQ. HAQ gives more reasonable policy to allocate the bits for each layer and pushes all the points to the upper right corner that is more efficient. 

\input{figures/haq/bismo_latency_transfer.tex}

Finally, we evaluate the transfer ability of our framework: first train our agent on one network (MobileNet-V1), then directly apply the agent to another network (MobileNet-V2) (see \tbl{tbl:bismo_transfer}). Our quantization policy transferred from MobileNet-V1 to MobileNet-V2 performs much better than the fixed-bitwidth baseline and is only slightly worse than the quantization policy directly searched for MobileNet-V2. This experiment validates that our RL agent generalizes well to different network architectures. That's to say, given a new model that the agent hasn't seen before, it can utilize its past knowledge to give a decent quantization policy, saving the design cycle.

%% file: figures/haq/teaser_tables.tex
\begin{table}[!t]
    \renewcommand*{\arraystretch}{1.4}
    \small\centering
    \begin{tabular}{lccc}
        \toprule
        & \multicolumn{3}{c}{Inference latency on} \\
        & \textbf{HW1} & \textbf{HW2} & \textbf{HW3} \\
        \midrule
        Best Q. policy for \textbf{HW1} & \cellcolor{red!15}\textbf{16.29} ms & 85.24 ms & 117.44 ms \\
        Best Q. policy for \textbf{HW2} & 19.95 ms & \cellcolor{red!15}\textbf{64.29} ms & 108.64 ms \\
        Best Q. policy for \textbf{HW3} & 19.94 ms & 66.15 ms & \cellcolor{red!15}\textbf{99.68} ms \\
        \bottomrule
    \end{tabular}
    \caption{Inference latency of MobileNet-V1~\cite{howard2017mobilenets} on three hardware architectures under different quantization policies. The quantization policy that is optimized for one hardware is not optimal for the other. This suggests we need a \textbf{specialized} quantization solution for different hardware architectures. (HW1: spatial accelerator\cite{sharma2018bit}, HW2: edge accelerator\cite{umuroglu2018bismo}, HW3: cloud accelerator\cite{umuroglu2018bismo}, batch = 16).}
    \label{tbl:teaser}
\end{table}

%% file: figures/haq/bismo_latency_v1_figures.tex
\begin{figure}[!t]
    \centering
    \includegraphics[width=1.0\linewidth]{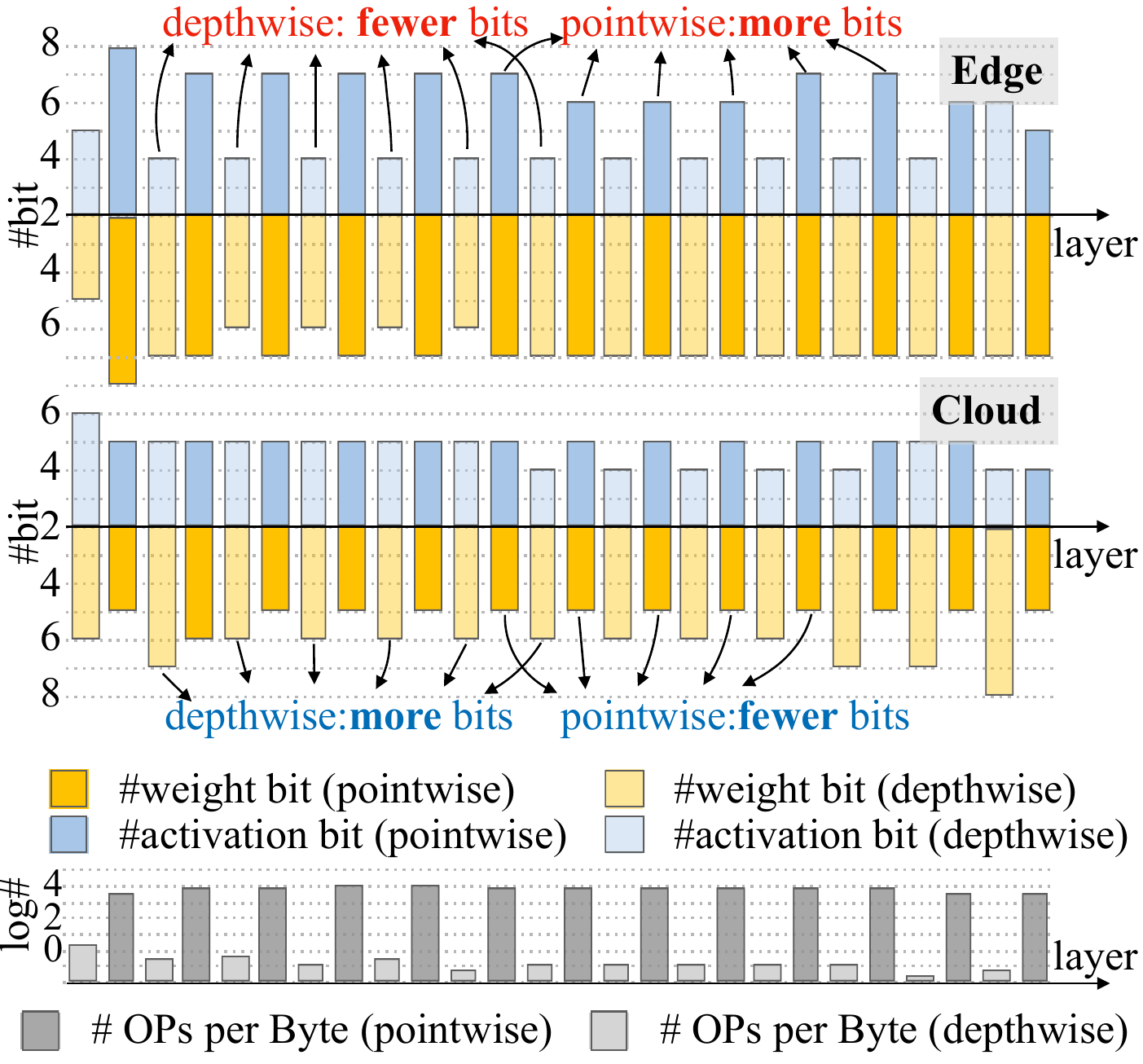}
    \caption{Quantization policy under latency constraints for MobileNet-V1.}
    \label{fig:bismo_v1}
\end{figure}

\begin{figure}[!t]
    \centering
    \includegraphics[width=\linewidth]{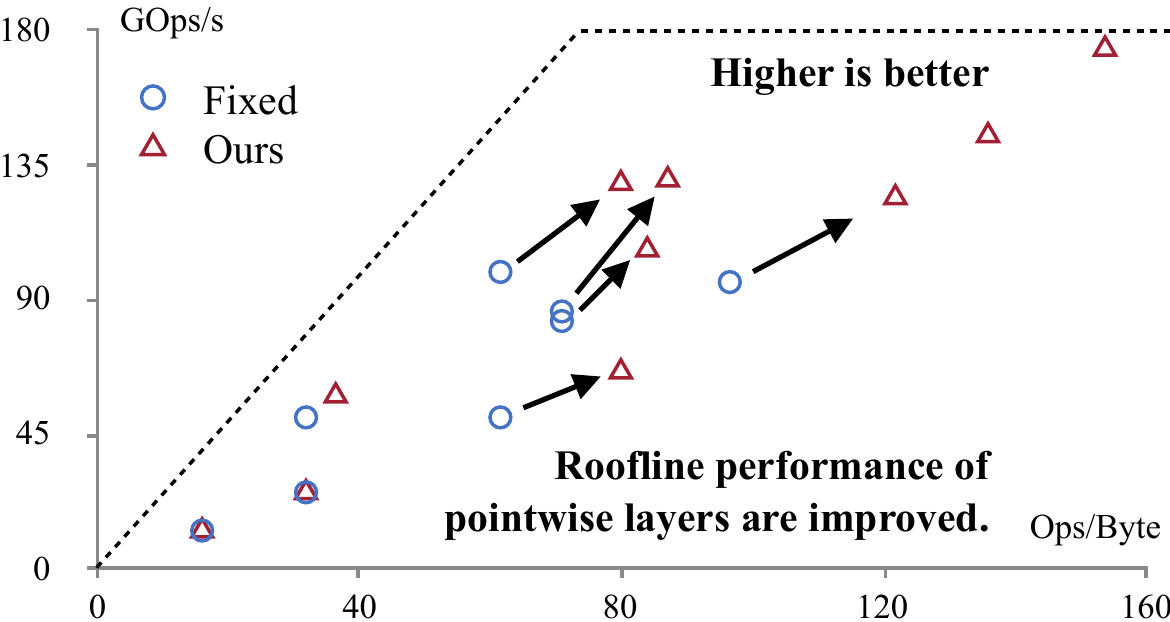}
    \caption{HAQ improves the roofline performance of pointwise layers in MobileNet-V1.}
    \label{fig:roofline}
\end{figure}

%% file: figures/haq/bismo_latency_tables.tex
\begin{table}[!t]
    \setlength{\tabcolsep}{5pt}
    \small\centering
    \begin{tabular}{lccccccccc} 
        \toprule
        & & \multicolumn{2}{c}{Edge Accelerator} & \multicolumn{2}{c}{Cloud Accelerator}  \\
        \cmidrule(lr){3-4}\cmidrule(lr){5-6}

        
        & Bitwidths & Top-1 & Latency & Top-1 & Latency  \\
        \midrule
        PACT & 4 bits & 62.44 & 45.45 ms & 61.39 & 52.15 ms \\
        Ours & \emph{flexible} & \textbf{67.40} & 45.51 ms & \textbf{66.99} & 52.12 ms \\
        \midrule
        PACT & 5 bits & 67.00 & 57.75 ms & 68.84 & 66.94 ms \\
        Ours & \emph{flexible} & \textbf{70.58} & 57.70 ms & \textbf{70.90} & 66.92 ms  \\
        \midrule
        PACT & 6 bits & 70.46 & 70.67 ms & 71.25 & 82.49 ms \\
        Ours & \emph{flexible} & \textbf{71.20} & 70.35 ms & \textbf{71.89} & 82.34 ms \\
        \midrule
        Original & 8 bits & 70.82 & 96.20 ms & 71.81 & 115.84 ms \\
        \bottomrule
    \end{tabular}
    \caption{Latency-constrained quantization on the edge and cloud accelerator on ImageNet.}
    \label{tbl:bismo_latency}
\end{table}

%% file: figures/haq/bismo_latency_transfer.tex
\begin{table}[!t]
    \renewcommand*{\arraystretch}{1.}
    \setlength{\tabcolsep}{6pt}
    \small\centering
    \begin{tabular}{lcccc}
        \toprule
        & Bitwidth & Top-1 & Latency \\
        \midrule
        PACT & 4 bits & 61.39 & 52.15 ms \\
        Ours (search for V2) & \emph{flexible} & 66.99 & 52.12 ms \\
        Ours (transfer from V1) & \emph{flexible} & 65.80 & 52.06 ms \\
        \midrule
        PACT & 5 bits & 68.84 & 66.94 ms \\
        Ours (search for V2) & \emph{flexible} & 70.90 & 66.92 ms \\
        Ours (transfer from V1) & \emph{flexible} & 69.90 & 66.93 ms \\
        \bottomrule
    \end{tabular}
    \caption{Our RL agent is able to generalize well to different network architectures: the quantization policy transferred from MobileNet-V1 to MobileNet-V2 performs very close to directly searching for MobileNet-V2. Both outperfomed the the fixed-bitwidth baseline.}
    \label{tbl:bismo_transfer}
\end{table}

%% file: contents/05_conclusion.tex
\section{Conclusion}

We present design automation techniques for efficient deep learning computing. There's plenty of room at the algorithm level to improve the hardware efficiency, but the large design space makes it difficult to be exhausted by human. We covered three aspects of design automation: specialized model design, compression and pruning, mixed-precision quantization. Such learning based design automation outperformed rule-base heuristics. Our framework reveals that the optimal design policies on different hardware architectures are drastically different, therefore specialization is important. We interpreted those design policies and believe the insights will inspire the future software and hardware co-design for efficient deep learning computing. 